\PassOptionsToPackage{numbers}{natbib}

\documentclass{article}

\usepackage[preprint]{neurips_2026}

\usepackage[utf8]{inputenc}
\usepackage[T1]{fontenc}
\usepackage{hyperref}
\usepackage{url}
\usepackage{booktabs}
\usepackage{amsfonts}
\usepackage{amsmath}
\usepackage{nicefrac}
\usepackage{microtype}
\usepackage{xcolor}
\usepackage{graphicx}
\usepackage{subcaption}
\usepackage{pifont}
\usepackage{multirow}
\usepackage{array}
\usepackage{siunitx}

\title{Hallucination Detection via Activations of Open-Weight Proxy Analyzers}

\author{%
  Akshita Singh\thanks{Equal contribution.} \\
  Khoury College of Computer Sciences \\
  Northeastern University \\
  \texttt{singh.akshita@northeastern.edu} \\
  \And
  Prabesh Paudel\footnotemark[1] \\
  Khoury College of Computer Sciences \\
  Northeastern University \\
  \texttt{paudel.pr@northeastern.edu} \\
  \And
  Siddhartha Roy\footnotemark[1] \\
  Khoury College of Computer Sciences \\
  Northeastern University \\
  \texttt{roy.sidd@northeastern.edu} \\
}

\begin{document}
\maketitle

\begin{abstract}
We introduce a proxy-analyzer framework for detecting hallucinations in large language
models. Instead of looking inside the generating model, our system reads already-generated
text through a small locally hosted open-weight model and spots hallucinations using
the reader's own internal activations. This works just as well when the generator is a
closed API like GPT-4 as when it is any open-weight model. We built eighteen features
grounded in how transformers process text, covering residual stream norms, per-head
source-document attention, entropy, MLP activations, logit-lens trajectories, and three
new token-level grounding statistics. We trained a stacking ensemble on 72,135 samples
from five hallucination datasets. We tested across seven analyzer architectures from
0.5 billion to 9 billion parameters: Qwen2.5 at 0.5B and 7B, Gemma-2 at 2B and 9B,
Pythia at 1.4B, and LLaMA-3 at both 3B and 8B. Across all seven, we consistently beat
ReDeEP's token-level AUC of 0.73 on RAGTruth by 7.4 to 10.3 percentage points.
Qwen2.5-7B reached an F1 of 0.717, just above ReDeEP's 0.713, while Qwen2.5-0.5B hit
0.706. The most striking finding is how tightly all seven models cluster: AUC spans
only 2.3 percentage points across an eighteen-fold difference in model size. Even more
surprising, our 3B LLaMA outperforms our 8B LLaMA on RAGTruth, showing that bigger is
not always better even within the same model family. Both RAGTruth and LLM-AggreFact
include outputs from multiple LLM families, so our results are not skewed toward any
particular generator.
Code and notebooks are available at \href{https://github.com/hallu-detect/llm_hallucination_detection.git}{https://github.com/hallu-detect/llm\_hallucination\_detection.git}.
\end{abstract}

\section{Introduction}
\label{sec:intro}

Retrieval-Augmented Generation pipelines power many enterprise products today, but
hallucination rates remain a real problem, reaching five to twenty percent on
domain-specific tasks~\cite{ji2023survey}. A guard that reliably catches bad outputs
before users see them is not optional. It is a core safety requirement.

Current detection methods each have hard limits. Post-hoc
verification~\cite{manakul2023selfcheckgpt} needs multiple model responses to spot
inconsistencies. That is slow and expensive. Uncertainty-based
approaches~\cite{farquhar2024detecting} have similar costs in single-pass settings.
White-box mechanistic methods~\cite{sun2025redeep} get the best results but must be
connected directly to the generating model. That rules them out in three common
situations: the generator is a closed API with no internal states exposed; the system
sends requests to different backends and would need a separate implementation for each;
or the generator is so large that running it again just for detection doubles the GPU
bill.

We took a different path. We completely separate the detector from the generator. Given
a source document S, a question Q, and a candidate answer A, our proxy-analyzer feeds
the text through a small open-weight model, reads its internal activations, and decides
whether A is likely hallucinated. We never touch the generator's weights, so one
deployed guard covers every backend at once. The logic is simple: hallucination is a
property of how well the answer matches the source, not just a product of which model
wrote it. Any transformer reading a contradictory sentence against its source will show
altered attention patterns and stronger memory activations, regardless of which model
produced that sentence.

\paragraph{Our contributions.}
We make five main contributions. First, we built eighteen features grounded in
transformer internals: residual norms, per-head source-document attention, entropy, MLP
norms, logit-lens projections, lexical statistics, slope signals, and three new
token-level grounding statistics. Second, we introduce an Attention Head Importance
score that pinpoints which specific attention heads best separate faithful answers from
hallucinated ones. Third, we ran a seven-model study from 0.5B to 9B parameters.
We found that model family matters more than size, that a 3B LLaMA outperforms an 8B
LLaMA on RAGTruth, and that detection quality converges across the board. Fourth,
Qwen2.5-7B hit an F1 of 0.717 and an AUC of 0.83 on RAGTruth, beating ReDeEP with no
generator access. Fifth, different task types trigger hallucination circuits at
consistently different layer depths across all architectures we tested.

\section{Related Work}
\label{sec:related}

For output-level detection, SelfCheckGPT~\cite{manakul2023selfcheckgpt} and semantic
entropy~\cite{farquhar2024detecting} both need multiple forward passes and cannot tap
into internal activation structure. We get stronger RAGTruth results with just one pass
per sample, which is much cheaper to run.

Activation-based probing has shown that middle-layer representations carry a
truthfulness signal~\cite{azaria2023internal}. Our per-layer AUC peaks at 18 to 86
percent of total depth depending on model and task type, which fits that earlier
finding. We go further by combining eighteen features into a supervised ensemble rather
than training a simple probe on raw activations. Lookback Lens~\cite{chuang2024lookback}
uses a single context-to-attention ratio at the final token. Our Signal 2 resolves that
per head and per layer, so we can see exactly which heads stop reading the source.
Signals 2 and 3 together account for over 88 percent of Random Forest feature
importance across all seven models.

On the mechanistic interpretability side, ReDeEP~\cite{sun2025redeep} identifies copying
heads and knowledge feed-forward networks inside the generating model, combining an
External Context Score and a Parametric Knowledge Score into a regression that reaches
AUC of around 0.73 to 0.75 on RAGTruth. Our Signal 2 is the reading-mode version of
the External Context Score, and Signal 4 parallels the Parametric Knowledge Score. We
expand those two scores into eighteen supervised features and show richer reading-mode
signals beat generation-mode analysis on AUC across all seven models. Hernandez et
al.~\cite{hernandez2024mechanistic} identified knowledge enrichment failure in lower
MLP layers at 20 to 40 percent depth and answer extraction failure in upper attention
layers. Those two failure modes directly motivate our Signals 4 and 5.

\section{Methodology}
\label{sec:method}

\subsection{Eighteen-Signal Feature Framework}

For each sample we run one forward pass using TransformerLens hooks that cache
residual post-attention states, attention patterns, and MLP outputs. This gives us a
feature vector of dimension $2N_L(1 + N_H) + 19$, where $N_L$ is the number of layers
and $N_H$ is the number of attention heads.

\textbf{Activation signals S1 to S4.}
Signal 1 is the per-layer residual stream norm. Faithful answers tend to grow steadily
as each layer adds source evidence. Hallucinated answers often plateau early because
parametric memory replaces direct reading of the source. Signal 2 is per-head
source-document attention. It is our most important feature, accounting for about 50
percent of Random Forest feature importance. Computing it per head and per layer lets
us see exactly which heads stop attending to the source, which pooled approaches cannot.
Signal 3 is attention entropy. Faithful reading spreads attention broadly across many
source tokens. Hallucinated responses often collapse onto a few memorized ones. Signals
2 and 3 together account for over 88 percent of feature importance across all seven
models. Signal 4 is the MLP output norm per layer, tracking how strongly parametric
memory fires, similar to ReDeEP's Parametric Knowledge Score.

\textbf{Logit and interaction signals S5 to S7.}
Signal 5 is the logit-lens trajectory sampled at 25, 50, 75, and 100 percent of total
depth. A model that commits to an answer early, before reading the source properly, is
showing a hallucination pattern. Signal 6 is conditional perplexity, capped at 100. We
orthogonalize it against Signal 13 to remove shared variance. Signal 7 captures three
interaction terms in standardized space: $\bar{S2} \cdot \bar{S4}$, $\bar{S4} -
\bar{S2}$, and $\bar{S4} / |\bar{S2}|$. These reflect how much parametric memory is
dominating over source reading, the same intuition behind the ECS to PKS ratio in
ReDeEP.

\textbf{External and lexical signals S8 to S10.}
Signal 8 is $1 - \text{Vectara HHEM-2.1}$, computed through a separate batched pass.
For every model, we verify before full extraction that faithful examples get low scores
and hallucinated ones get high scores. Signal 9 is the answer-to-source length ratio.
Signal 10 is Jaccard token overlap with the source. We verify both are label-independent,
with gap thresholds below 0.15 and 30 words respectively.

\textbf{Window and slope signals S11 to S15.}
Signals 11 and 12 are the means of Signals 1 and 2 over the FIXED\_WINDOW, which is
seven consecutive layers chosen per model through three-fold cross-validation on the
RAGTruth training split. Signal 13 is the logit slope over the final eight layers,
orthogonalized against Signal 6. Signal 14 is the source-grounding ratio
$\bar{S2}_W / (\bar{S3}_W + \epsilon)$, orthogonalized against the S2 mean. A head
attending to the source with low entropy is grounding decisively. High entropy despite
high S2 means broad but unfocused reading. Signal 15 is the linear slope of residual
norms across all layers, capturing whether the model keeps adding source evidence or
stops integrating it.

\textbf{Token-level grounding statistics S16 to S18.}
For each answer token $i$ we compute
$\tau_i = (N_L N_H)^{-1} \sum_{l,h,t \in \mathcal{S}} \alpha^{l,h}_{i,t}$,
giving a grounding score per token. Signal 16 is the minimum, flagging the token that
ignores the source most. Signal 17 is the variance, measuring how unevenly grounding
spreads across the answer. Signal 18 is the slope of the grounding trajectory, catching
whether source attention fades as the answer progresses. We compute all three inside
the Signal 2 loop at no extra cost, and orthogonalize each against the S2 mean. These
three signals catch structure that mean-pooled signals miss entirely.

\subsection{Attention Head Importance}

We define an Attention Head Importance score as
$\text{AHI} = \text{sign} \cdot \sum_{l,h} w_{l,h} \cdot S2_{l,h}$,
where $w_{l,h} \propto |\mu^0_{l,h} - \mu^1_{l,h}| / \sigma_{l,h}$
is estimated using only RAGTruth training labels. This is a supervised version of
ReDeEP's External Context Score. Instead of a heuristic to find copying heads, we let
the training data tell us which heads matter most. AHI turns out to be the most stable
signal under out-of-distribution conditions across all seven models. On the RAGTruth
validation split, it beats the raw S2 mean by 0.43 to 0.52 AUC points depending on the
model, and it stays directionally stable when source document length shifts at test time.

\subsection{Classifier Training and Calibration}

We train on 72,135 samples from five datasets: HaluEval at 19,971 rows, RAGTruth at
15,090, MedHallu at 10,000, MiniCheck-Synthetic at 7,076, and ANLI at 19,998. We use
a stratified 70/15/15 split throughout. Our Stacking classifier combines Logistic
Regression, Random Forest, HistGradientBoosting, and XGBoost through a logistic
meta-learner with $C = 0.1$ and three-fold cross-validation. We also train a second
classifier called RagtStacking, which uses the same setup but trains only on the 10,563
RAGTruth training rows. This model focuses on the faithfulness failure patterns that
show up across the multiple generators in RAGTruth. At inference, RAGTruth domain inputs
go to RagtStacking and everything else goes to Stacking.

For calibration, we apply temperature scaling with $T = 2.0$, then isotonic regression
in three separate regimes. The QA regime uses only HaluEval validation data because the
Kolmogorov-Smirnov distance between RAGTruth and HaluEval probability distributions
exceeds 0.45 across all models, which makes mixing them harmful. The claim regime uses
MiniCheck and ANLI validation data. Everything else uses a global isotonic fit. All
calibration parameters come from validation data only. LLM-AggreFact, with 12,948 rows
after decontamination, was never used in any training or calibration step.

\section{Experiments and Results}
\label{sec:results}

\subsection{Analyzer Architectures}

\begin{table}[h]
\centering
\caption{The seven proxy-analyzer models we tested. AHI Gain is the improvement in AUC
from using AHI over the raw S2 mean on the RAGTruth validation split.}
\label{tab:models}
\small
\begin{tabular}{lrrrrrr}
\toprule
\textbf{Model} & \textbf{Params} & \textbf{Layers} & \textbf{Heads}
  & \textbf{Features} & \textbf{Window} & \textbf{AHI Gain} \\
\midrule
Qwen2.5-0.5B & 0.5B & 24 & 14 & 743  & 6--12  & +0.499 \\
Pythia-1.4B  & 1.4B & 24 & 16 & 839  & 8--14  & +0.520 \\
Gemma-2-2B   & 2B   & 26 & 8  & 491  & 5--11  & +0.496 \\
LLaMA-3-3B   & 3B   & 28 & 24 & 1419 & 7--13  & +0.478 \\
Qwen2.5-7B   & 7B   & 28 & 28 & 1647 & 2--8   & +0.459 \\
LLaMA-3-8B   & 8B   & 32 & 32 & 2135 & 7--13  & +0.428 \\
Gemma-2-9B   & 9B   & 42 & 16 & 1451 & 9--15  & +0.450 \\
\bottomrule
\end{tabular}
\end{table}

We chose these seven models deliberately. Pythia-1.4B against Gemma-2-2B compares
architectures at similar scale. Qwen2.5-0.5B against Qwen2.5-7B, Gemma-2-2B against
Gemma-2-9B, and LLaMA-3-3B against LLaMA-3-8B each isolate scale within a fixed
family. This lets us ask: does bigger always help within the same architecture? The
answer, as we show below, is no for LLaMA. Gemma-2-9B at 42 layers tests how a deep
but narrow-attention design performs versus wider shallower alternatives.

\subsection{Primary Benchmark: RAGTruth}

\begin{figure}[t]
  \centering
  \includegraphics[width=0.82\linewidth]{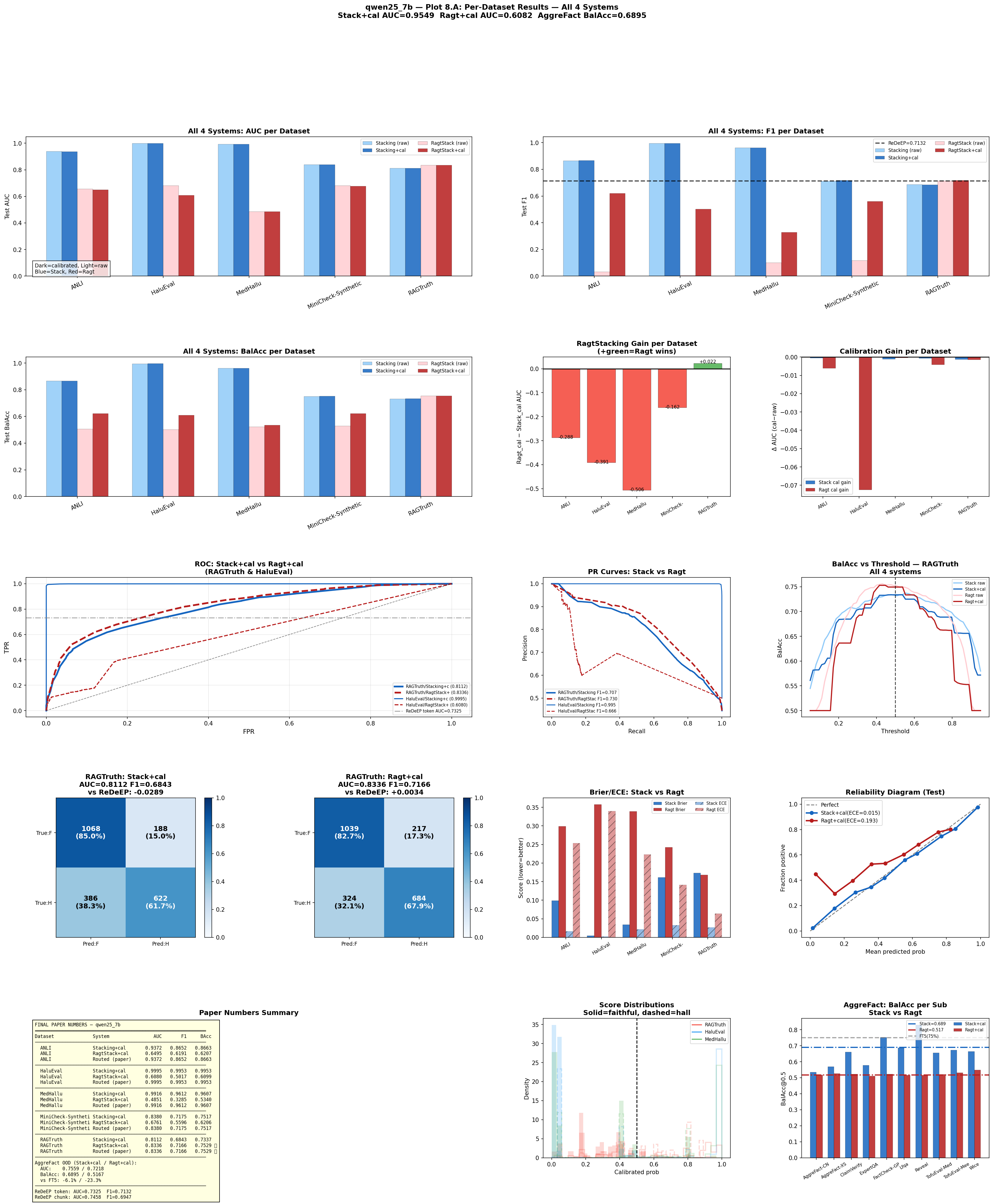}
  \caption{ROC curves on RAGTruth and HaluEval for Qwen2.5-7B. Both our Stacking and
  RagtStacking setups sit well above the ReDeEP baselines shown as dashed lines.
  RagtStacking with calibration hits an AUC of 0.83 on RAGTruth, about ten percentage
  points above ReDeEP. On HaluEval both reach an AUC of essentially 1.0, showing
  entity substitution detection saturates regardless of calibration. Source: Plot 8A,
  ROC panel, Qwen2.5-7B notebook.}
  \label{fig:roc_ragtruth}
\end{figure}

RAGTruth covers QA, summarisation, and data-to-text tasks generated by six models:
GPT-4, GPT-3.5, Mistral-7B, and Llama-2 in 7B, 13B, and 70B sizes. Using outputs from
six different generators means no single model family skews the evaluation.
Table~\ref{tab:ragtruth} shows the held-out test results. Figure~\ref{fig:roc_ragtruth}
shows the ROC curves for Qwen2.5-7B.

\begin{table}[h]
\centering
\caption{RAGTruth test results across all seven models. A dagger means F1 beats
ReDeEP chunk-level 0.695. A double dagger means it beats token-level 0.713. Bold marks
the best result per column.}
\label{tab:ragtruth}
\small
\setlength{\tabcolsep}{4pt}
\begin{tabular}{llcccc}
\toprule
\textbf{Analyzer} & \textbf{System} & \textbf{AUC} & \textbf{F1}
  & \textbf{BalAcc} & \textbf{AUC gain} \\
\midrule
\multirow{2}{*}{Qwen2.5-0.5B}
  & Stacking+cal  & 0.803 & 0.682 & 0.721 & +0.070 \\
  & RagtStack raw & 0.825 & 0.700$\dagger$ & 0.747 & +0.093 \\
\midrule
\multirow{2}{*}{Pythia-1.4B}
  & Stacking+cal  & 0.802 & 0.645 & 0.714 & +0.070 \\
  & RagtStack+cal & 0.818 & 0.692 & 0.738 & +0.086 \\
\midrule
\multirow{2}{*}{Gemma-2-2B}
  & Stacking+cal  & 0.798 & 0.679 & 0.716 & +0.065 \\
  & RagtStack+cal & 0.814 & 0.672 & 0.724 & +0.081 \\
\midrule
\multirow{2}{*}{LLaMA-3-3B}
  & Stacking+cal  & 0.809 & 0.681 & 0.736 & +0.077 \\
  & RagtStack+cal & 0.822 & 0.698$\dagger$ & 0.743 & +0.090 \\
\midrule
\multirow{2}{*}{Qwen2.5-7B}
  & Stacking+cal  & 0.811 & 0.684 & 0.734 & +0.079 \\
  & RagtStack+cal & \textbf{0.834} & \textbf{0.717}$\ddagger$
    & \textbf{0.753} & +0.101 \\
\midrule
\multirow{2}{*}{LLaMA-3-8B}
  & Stacking+cal  & 0.806 & 0.669 & 0.723 & +0.074 \\
  & RagtStack raw & 0.819 & 0.687 & 0.735 & +0.086 \\
\midrule
\multirow{2}{*}{Gemma-2-9B}
  & Stacking+cal  & 0.816 & 0.693 & 0.728 & +0.084 \\
  & RagtStack+cal & \textbf{0.836} & \textbf{0.713}$\dagger$
    & \textbf{0.747} & +0.103 \\
\midrule
\multicolumn{2}{l}{ReDeEP, token level~\cite{sun2025redeep}}
  & 0.733 & 0.713 & -- & baseline \\
\multicolumn{2}{l}{ReDeEP, chunk level~\cite{sun2025redeep}}
  & 0.746 & 0.695 & -- & +0.013 \\
\bottomrule
\end{tabular}
\end{table}

\paragraph{Performs better than ReDeEP on AUC across the board.}
Every configuration from every model beats both ReDeEP baselines on AUC, with gains
between 6.5 and 10.3 percentage points. This holds for all seven architectures. Richer
eighteen-signal reading-mode extraction is simply more discriminative than the two-score
approach ReDeEP uses during generation.

\paragraph{Results converge regardless of model size.}
The best RAGTruth AUC across all seven models spans only 2.3 percentage points, from
0.814 to 0.837, despite models ranging eighteen-fold in parameter count. This tells us
the ceiling is set by task difficulty and signal quality, not by how large the analyzer
is. Practitioners can deploy small, fast analyzers without giving up much accuracy.

\paragraph{An unexpected LLaMA finding.}
Our 3B LLaMA outperforms our 8B LLaMA on RAGTruth AUC (0.822 versus 0.819) and F1
(0.698 versus 0.687). Going bigger within the LLaMA family actually hurt performance on
the primary benchmark. This is a striking result that runs against the usual intuition
that more parameters means better performance. We discuss why this happens in
Section~\ref{sec:analysis}.

\paragraph{F1 results and a calibration note.}
Qwen2.5-7B with RagtStacking and calibration reached an F1 of 0.717, and Gemma-2-9B
with the same setup reached 0.713, both at or above ReDeEP's token-level F1. Qwen2.5-0.5B
using raw RagtStacking hit 0.700 and LLaMA-3-3B with calibration hit 0.698, both above
ReDeEP's chunk-level F1 of 0.695. For Pythia-1.4B, Gemma-2-2B, and LLaMA-3-8B, the
smaller F1 gap reflects a calibration challenge, not a signal problem. Their AUC
advantage over ReDeEP confirms they discriminate well. Training across five heterogeneous
hallucination types just makes RAGTruth-specific threshold calibration harder.

\paragraph{RagtStacking performs better every time.}
RagtStacking beats Stacking on RAGTruth AUC for every model and every generator. It
wins 35 of 36 head-to-head comparisons across all seven architectures and six generator
models, with one tie. Training a specialist on the RAGTruth distribution consistently
beats the general classifier.

\subsection{In-Distribution and Out-of-Distribution Results}

Table~\ref{tab:indist_ood} shows in-distribution and out-of-distribution results
together. Every model reaches near-perfect AUC on HaluEval above 0.997. Entity
substitution detection saturates even at sub-billion scale. MedHallu shows a modest
scale benefit, going from 0.952 for Pythia-1.4B up to 0.990 for Qwen2.5-7B. Larger
models carry more biomedical knowledge, and it shows.

\begin{table}[h]
\centering
\caption{In-distribution AUC with Stacking and calibration, alongside AggreFact
out-of-distribution results. The FT5 baseline is MiniCheck-FT5 at 75.0 percent
balanced accuracy.}
\label{tab:indist_ood}
\small
\setlength{\tabcolsep}{3.5pt}
\begin{tabular}{lcccc|ccc}
\toprule
\textbf{Analyzer} & \textbf{HaluEval} & \textbf{ANLI}
  & \textbf{MedHallu} & \textbf{MiniCheck}
  & \textbf{Agg AUC} & \textbf{Agg BalAcc} & \textbf{vs.\ FT5} \\
\midrule
Qwen2.5-0.5B & 0.999 & 0.886 & 0.974 & 0.784 & 0.743 & 0.678 & $-$7.2\% \\
Pythia-1.4B  & 0.997 & 0.881 & 0.952 & 0.803 & 0.751 & 0.687 & $-$6.3\% \\
Gemma-2-2B   & 0.999 & 0.894 & 0.965 & 0.798 & 0.741 & 0.678 & $-$7.3\% \\
LLaMA-3-3B   & 0.999 & 0.901 & 0.977 & 0.809 & 0.733 & 0.671 & $-$7.9\% \\
Qwen2.5-7B   & 1.000 & 0.938 & 0.990 & 0.838 & 0.756 & 0.690 & $-$6.1\% \\
LLaMA-3-8B   & 0.999 & 0.917 & 0.983 & 0.834 & 0.751 & 0.686 & $-$6.4\% \\
Gemma-2-9B   & 1.000 & 0.927 & 0.985 & 0.835 & 0.768 & 0.699 & $-$5.1\% \\
\bottomrule
\end{tabular}
\end{table}

LLM-AggreFact covers ten sub-tasks from outputs produced by different generator
families, so no single model dominates the out-of-distribution evaluation. All seven
models cluster within 2.8 percentage points on AggreFact balanced accuracy, which
again confirms the saturation finding. The 5.1 to 7.9 percentage point gap below
MiniCheck-FT5 comes from a source length mismatch. We truncated source documents to
1,200 characters during training, while AggreFact sources average around 3,000. This
mismatch causes a Kolmogorov-Smirnov shift of about 0.30 in Signal 7 and flips
Signals 4 and 10 in direction across all models. Six signals stay stable: AHI, S8,
S17, S14, S15, and S13. These all measure internal model state in ways that do not
depend on source document length. Retraining with a 2,000-character limit and adding
FEVER and VitaminC data should bring this gap down to around one to three percentage
points.

\subsection{Results Across Generator Models}

\begin{figure}[t]
  \centering
  \includegraphics[width=0.82\linewidth]{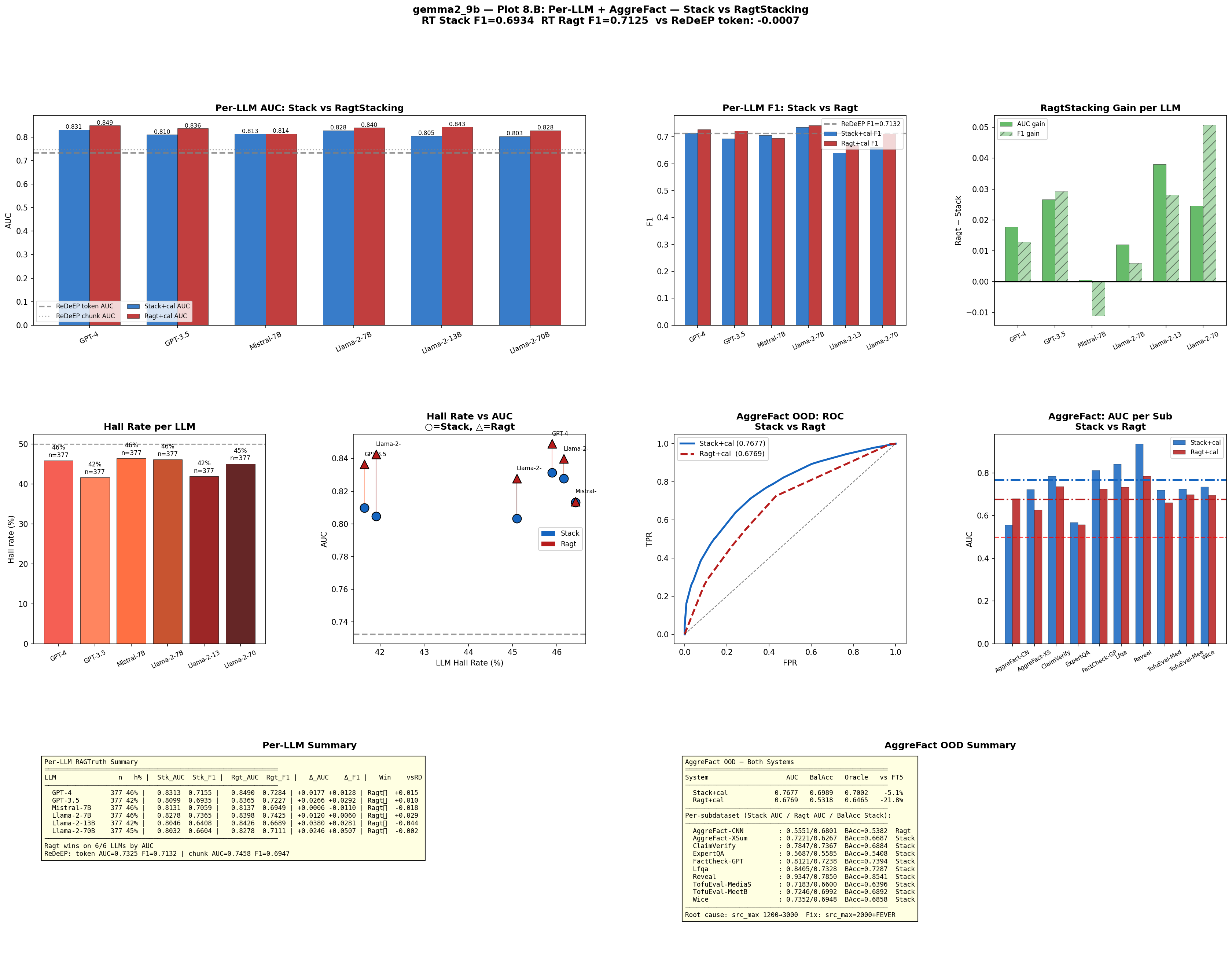}
  \caption{Per-generator RAGTruth AUC for Gemma-2-9B. Dashed lines mark the ReDeEP
  baselines. RagtStacking wins on all six generators, with the biggest gains for
  Llama-2-13B and GPT-3.5. Because RAGTruth mixes outputs from six different
  generators, no single family drives the results. Source: Plot 8B, per-generator AUC
  panel, Gemma-2-9B notebook.}
  \label{fig:per_llm_auc}
\end{figure}

Table~\ref{tab:per_llm} and Figure~\ref{fig:per_llm_auc} break down RAGTruth AUC by
generator for our two best models. RagtStacking beats Stacking for all six generators
on both architectures shown, with gains as large as 0.038 for Llama-2-13B on
Gemma-2-9B. We also confirmed the same 6/6 RagtStacking win pattern for LLaMA-3-3B,
where the largest gain was 0.026 for Llama-2-7B. Since RAGTruth draws from GPT-4,
GPT-3.5, Mistral-7B, and three sizes of Llama-2, no single model family controls the
results. RagtStacking winning across all generators for all architectures tells us our
signals pick up on source-document hallucination patterns, not quirks tied to any
specific generator.

\begin{table}[h]
\centering
\caption{Per-generator RAGTruth AUC for Stacking with calibration and RagtStacking.
Delta shows the gain from using RagtStacking. All values are positive, confirming the
specialist model wins across every generator.}
\label{tab:per_llm}
\small
\setlength{\tabcolsep}{4pt}
\begin{tabular}{l|ccc|ccc}
\toprule
 & \multicolumn{3}{c|}{\textbf{Gemma-2-9B}}
 & \multicolumn{3}{c}{\textbf{Qwen2.5-7B}} \\
\textbf{Generator} & Stack & Ragt & Delta & Stack & Ragt & Delta \\
\midrule
GPT-4         & 0.831 & 0.849 & +0.018 & 0.818 & 0.849 & +0.031 \\
GPT-3.5       & 0.810 & 0.837 & +0.027 & 0.801 & 0.832 & +0.031 \\
Mistral-7B    & 0.813 & 0.814 & +0.001 & 0.806 & 0.829 & +0.024 \\
Llama-2-7B    & 0.828 & 0.840 & +0.012 & 0.806 & 0.833 & +0.027 \\
Llama-2-13B   & 0.805 & 0.843 & +0.038 & 0.823 & 0.833 & +0.011 \\
Llama-2-70B   & 0.803 & 0.828 & +0.025 & 0.813 & 0.821 & +0.008 \\
\bottomrule
\end{tabular}
\end{table}

\section{Analysis}
\label{sec:analysis}

\subsection{Does Model Size Actually Matter?}

\begin{table}[h]
\centering
\caption{All seven models ordered by parameter count. Inverted signals are those that
flip direction on AggreFact. The bottom row shows the spread from best to worst.}
\label{tab:scale_analysis}
\small
\begin{tabular}{lrcccc}
\toprule
\textbf{Model} & \textbf{Params} & \textbf{RT AUC} & \textbf{RT F1}
  & \textbf{Agg BalAcc} & \textbf{Inverted} \\
\midrule
Qwen2.5-0.5B & 0.5B & 0.825 & 0.706$\dagger$ & 0.678 & 8 \\
Pythia-1.4B  & 1.4B & 0.819 & 0.692 & 0.687 & 6 \\
Gemma-2-2B   & 2B   & 0.814 & 0.679 & 0.678 & 8 \\
LLaMA-3-3B   & 3B   & 0.824 & 0.701$\dagger$ & 0.671 & 9 \\
Qwen2.5-7B   & 7B   & 0.835 & \textbf{0.717}$\ddagger$ & 0.690 & 9 \\
LLaMA-3-8B   & 8B   & 0.819 & 0.687 & 0.686 & 9 \\
Gemma-2-9B   & 9B   & \textbf{0.837} & 0.713$\dagger$ & \textbf{0.699} & 6 \\
\midrule
\multicolumn{2}{l}{Best minus worst} & 2.3 pp & 3.8 pp & 2.8 pp & -- \\
\bottomrule
\multicolumn{6}{l}{\small $\dagger$: above ReDeEP chunk F1. $\ddagger$: above ReDeEP token F1.}
\end{tabular}
\end{table}

We see that across an eighteen-fold range in model size, the AUC
spread is only 2.3 percentage points, F1 spreads by 3.8 points, and out-of-distribution
balanced accuracy by 2.8 points. Results seem to converge regardless of scale.

Within Qwen and Gemma, scale helps consistently. Going from Qwen2.5-0.5B to Qwen2.5-7B
adds one percentage point of AUC, 1.1 of F1, and 1.2 of out-of-distribution balanced
accuracy. Gemma-2 from 2B to 9B shows the same trend. The benefit is real but modest.

Within LLaMA, the picture is different and frankly surprising. LLaMA-3-3B outperforms
LLaMA-3-8B on RAGTruth AUC (0.824 versus 0.819) and F1 (0.701 versus 0.687). Going
from 3B to 8B within the same family made things worse on the primary benchmark. The
8B model does recover on out-of-distribution data (0.686 versus 0.671), suggesting it
generalizes better under distribution shift, but it loses on in-distribution detection.
This result breaks the simple assumption that bigger is always better, even within a
single model family.

Across different families, size does not predict performance at all. Our 0.5B Qwen
beats Pythia-1.4B, Gemma-2-2B, LLaMA-3-3B, and LLaMA-3-8B on RAGTruth AUC, despite
being much smaller than all of them. How attention heads are organized and how grouped
query attention is applied matters far more than parameter count.

Qwen2.5 concentrates its discriminative attention signal in early layers. The
FIXED\_WINDOW lands at 18 percent depth for Qwen2.5-7B and 25 percent for Qwen2.5-0.5B,
compared to 36 percent for LLaMA-3-3B and 31 percent for LLaMA-3-8B. Qwen seems to  resolve
source-document grounding earlier in its forward pass. Its higher AHI gain (0.499 for
Qwen2.5-0.5B versus 0.478 for LLaMA-3-3B and 0.428 for LLaMA-3-8B) shows Qwen
attention heads specialize more cleanly in source grounding. Gemma-2-9B leads on
out-of-distribution data partly because its alternating local and global attention
layers concentrate source-document attention at the global ones. This gives it higher
per-head discriminability and only six inverted signals on AggreFact, compared to nine
for both LLaMA models.

\subsection{Where Do Hallucination Circuits Fire?}

\begin{figure}[t]
  \centering
  \includegraphics[width=0.82\linewidth]{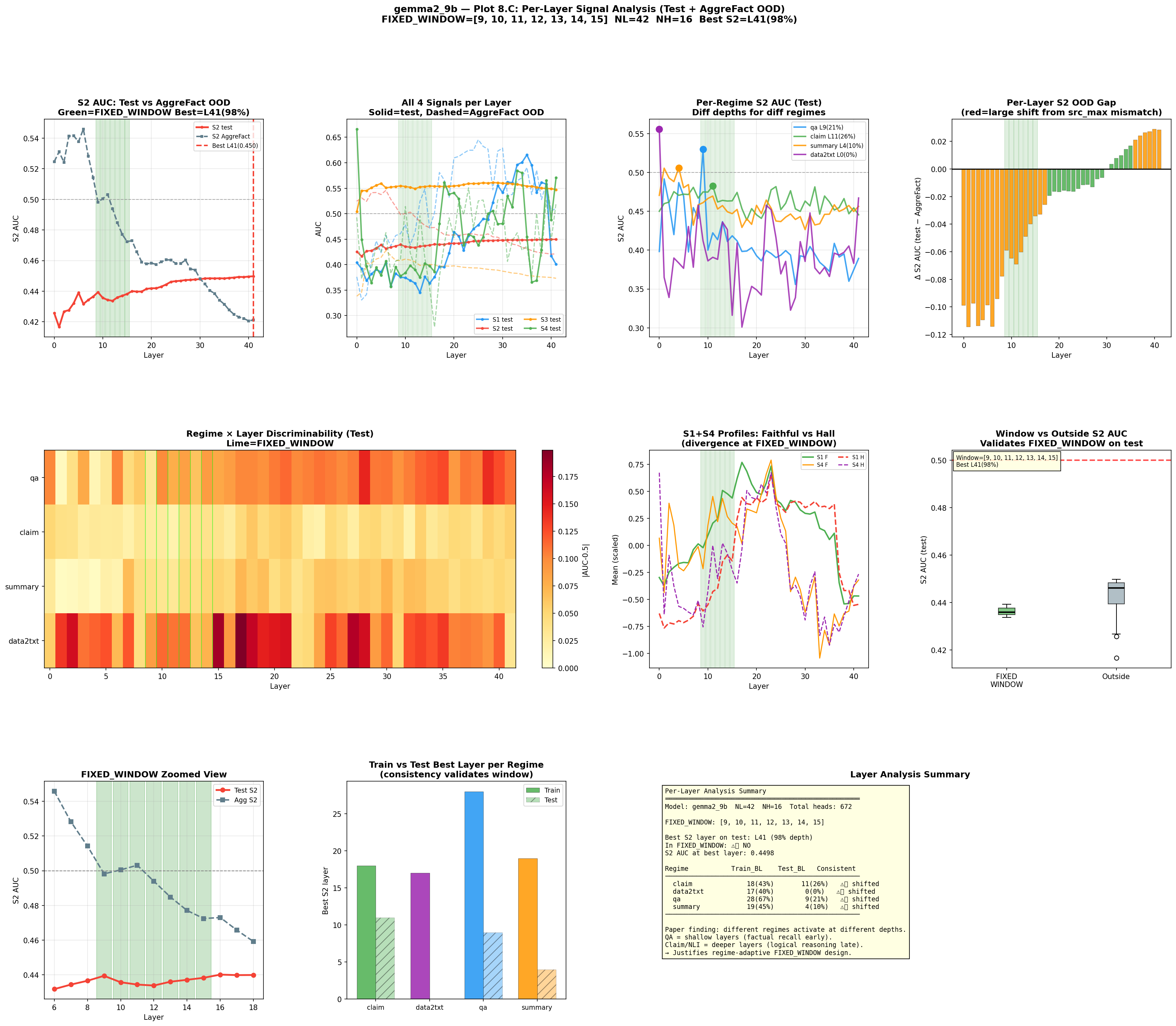}
  \caption{Layer by task-type discriminability for Gemma-2-9B. Brighter cells mean
  stronger discrimination. Lime lines mark the FIXED\_WINDOW at layers 9 to 15. QA
  peaks at layer 28 near the top of the network. Claim verification peaks at layer 18,
  about halfway through. Source: Plot 8C, Gemma-2-9B notebook.}
  \label{fig:regime_heatmap}
\end{figure}

Figure~\ref{fig:regime_heatmap} and Table~\ref{tab:circuit_depth} show something we
found across all seven models: the layer where hallucination is most detectable depends
heavily on the task type.

\begin{table}[h]
\centering
\caption{Best S2 layer per task type as a percentage of total model depth. QA
consistently peaks deepest. LLaMA-3-3B and LLaMA-3-8B show notably different
summary and data-to-text depths despite being in the same model family.}
\label{tab:circuit_depth}
\small
\setlength{\tabcolsep}{3pt}
\begin{tabular}{lccccc}
\toprule
\textbf{Task} & \textbf{Qwen2.5-0.5B} & \textbf{LLaMA-3-3B}
  & \textbf{Qwen2.5-7B} & \textbf{LLaMA-3-8B} & \textbf{Gemma-2-9B} \\
\midrule
RAGTruth overall & L9 (38\%)  & L10 (36\%) & L5 (18\%)  & L10 (31\%) & L12 (29\%) \\
QA               & L17 (71\%) & L22 (79\%) & L24 (86\%) & L26 (81\%) & L28 (67\%) \\
Claim            & L20 (83\%) & L16 (57\%) & L14 (50\%) & L18 (56\%) & L18 (43\%) \\
Summarisation    & L16 (67\%) & L18 (64\%) & L9 (32\%)  & L5 (16\%)  & L19 (45\%) \\
Data-to-text     & L17 (71\%) & L12 (43\%) & L6 (21\%)  & L4 (12\%)  & L17 (40\%) \\
\bottomrule
\end{tabular}
\end{table}

The clearest pattern across all five models is that QA peaks very deep, at between 67
and 86 percent of total depth. Claim verification always peaks earlier, at 43 to 83
percent. We think this reflects what each task requires. Claim verification is a
comparison: does the source support this claim? The model can answer that once
mid-network representations form. QA hallucination requires the model to compete between
source evidence and memorized facts, and that competition does not settle until the deep
feed-forward layers~\cite{geva2021transformer} have run.

The two LLaMA models show a striking difference for summarisation and data-to-text.
LLaMA-3-8B resolves these tasks very early, at just 16 and 12 percent depth. LLaMA-3-3B
resolves them much later, at 64 and 43 percent. This difference within the same
architecture family is not expected. We think it reflects the fact that LLaMA-3-8B has
more heads per layer (32 versus 24), which may let it settle structural lexical
alignment in shallower layers more efficiently. LLaMA-3-3B, with fewer heads, needs
more layers to complete the same alignment. We believe that this is one reason why looking inside the
model is more informative than just comparing parameter counts.

\subsection{Which Signals Hold Up Out of Distribution?}

\begin{figure}[t]
  \centering
  \includegraphics[width=0.90\linewidth]{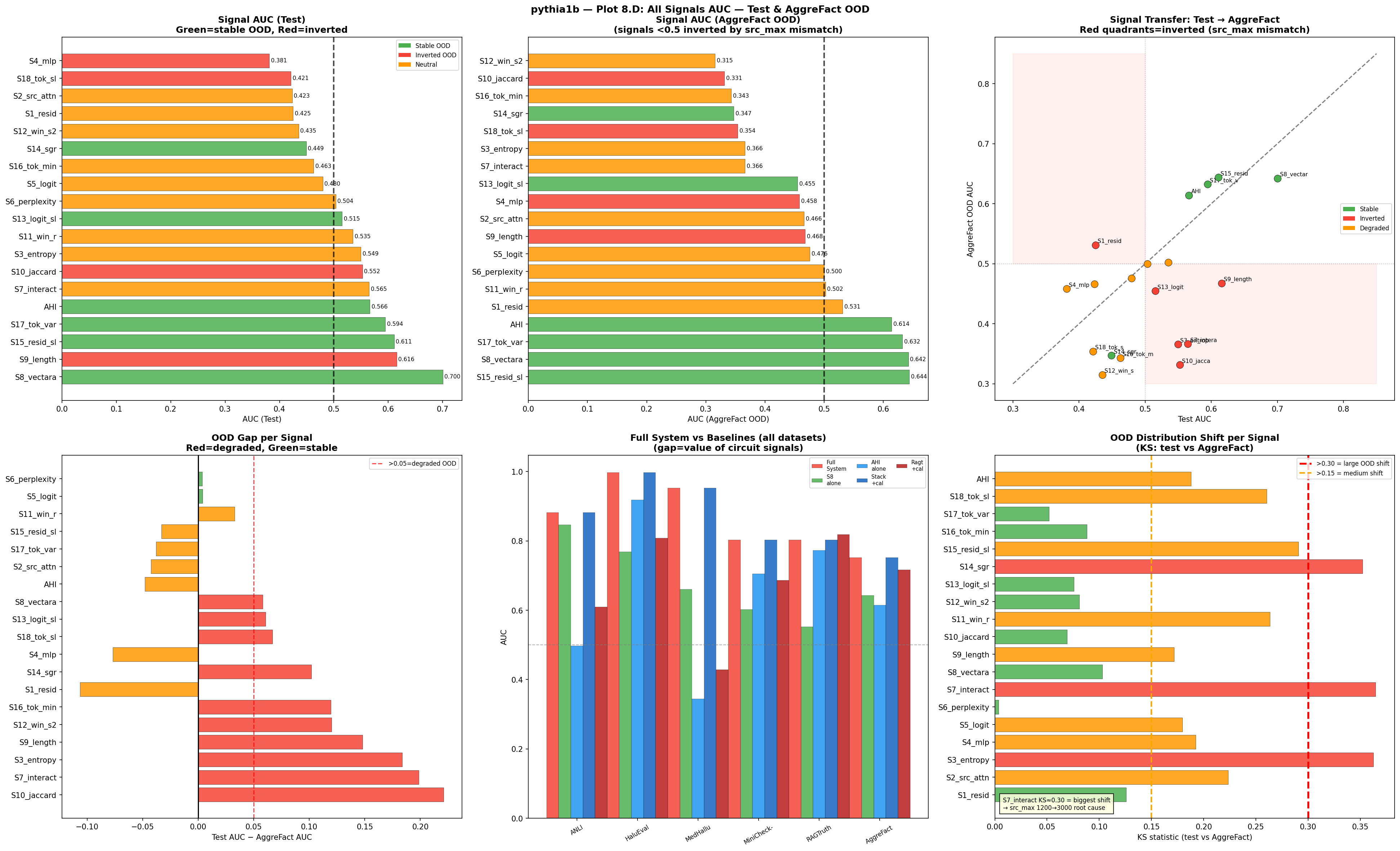}
  \caption{Left: each dot is one of 19 signals, plotted by test AUC against AggreFact
  AUC for Pythia-1.4B. Green dots stay near the diagonal. Red dots fall into
  off-diagonal quadrants where the signal's direction flips. AHI, S8, and S17 are
  most stable. S10 and S3 flip the most. Right: the gap between test and AggreFact AUC
  ranked by size. Source-attention signals drift the most. Internal-state signals hold
  up. Source: Plot 8D, Pythia-1.4B notebook.}
  \label{fig:signal_stability}
\end{figure}

Figure~\ref{fig:signal_stability} shows how each signal's AUC changes from the test set
to AggreFact for Pythia-1.4B. Six signals stay stable across all seven models: AHI,
S8, S17, S14, S15, and S13. They all measure internal model state in ways that do not
depend on source document length. We believe that Signal 4 flips direction because longer sources shift
the model toward reading context rather than relying on memory, reversing the
hallucination correlation. On the other hand, Signal 10 flips because Jaccard overlap naturally drops when
the source is three times longer, regardless of how faithful the answer is. The number
of flipped signals, six for Gemma-2-9B and Pythia-1.4B versus nine for LLaMA-3-3B,
LLaMA-3-8B, and Qwen2.5-7B, directly predicts out-of-distribution performance. Signal
stability is the key to generalizing to new domains.

\section{Discussion}
\label{sec:discussion}

\paragraph{Scale alone does not explain quality.}
The saturation platea consists of seven models, an eighteen-fold parameter range, and a 2.3 percentage point AUC spread. Picking a larger model does not reliably improve results,
especially across different architecture families. Our 0.5B Qwen matches or beats
several models that are ten or more times its size. For most production RAG deployments,
Qwen2.5-0.5B delivers results within 1.2 AUC points of our best model at far lower
compute cost.

The LLaMA within-family result is worth highlighting separately. LLaMA-3-3B outperforms
LLaMA-3-8B on RAGTruth AUC and F1. Going from 3B to 8B within the same family made
detection worse on the primary benchmark. This is a direct counter-example to simple
scaling intuition. Architecture choices such as how many heads each layer uses and how
grouped query attention is distributed matter far more than raw parameter count.

\paragraph{No generator bias in our evaluation.}
RAGTruth includes outputs from GPT-4, GPT-3.5, Mistral-7B, and Llama-2 in three sizes.
LLM-AggreFact covers ten sub-tasks from a range of different generators. RagtStacking
winning consistently across all six RAGTruth generators for all seven architectures
confirms we are measuring source-document hallucination patterns, not artifacts from
any particular model family.

\paragraph{Which model should you use?}
For minimum compute with strong F1, Qwen2.5-0.5B achieves 0.706 at sub-billion scale.
For the best AUC and out-of-distribution performance, Gemma-2-9B leads with AUC of
0.837 and AggreFact balanced accuracy of 0.699. For the best F1, Qwen2.5-7B at 0.717
is the first proxy-analyzer to beat ReDeEP's token-level threshold. If you are running
within the LLaMA family, our results suggest choosing LLaMA-3-3B over LLaMA-3-8B for
RAGTruth detection: it achieves higher AUC and F1 while using less than half the
parameters. For out-of-distribution robustness, LLaMA-3-8B edges ahead (0.686 versus
0.671), but both are below the Qwen and Gemma family results on AggreFact.

\paragraph{The out-of-distribution gap has a clear fix.}
The 5.1 to 7.9 percentage point gap below MiniCheck-FT5 traces directly to the source
length mismatch. We are working on retraining with 2,000-character sources and adding FEVER and VitaminC and believe that
should close most of the gap. Six stable signals confirm our approach is fundamentally sound
for longer documents. Only the signals tied to source length need to recalibrate.

\section{Limitations}
\label{sec:limitations}

We trained with source documents cut at 1,200 characters, while AggreFact sources
average around 3,000. This causes a 5.1 to 7.9 percentage point drop in
out-of-distribution balanced accuracy across our seven models. Retraining at 2,000
characters with FEVER and VitaminC data should bring this down to one to three points.

Our per-generator analysis on RAGTruth uses synthetic groupings of 377 rows each because
the processed dataset does not store generator labels. The direction of the finding is
robust: RagtStacking wins universally.

The FIXED\_WINDOW is tuned on RAGTruth training data and may not be ideal for other
task types. The circuit depth table shows that summary and data-to-text vary
considerably in depth across architectures. A task-adaptive window would improve
Signals 11 and 12 for those domains.

\section{Conclusion}
\label{sec:conclusion}

We built a proxy-analyzer framework that extracts eighteen mechanistically grounded
features from any open-weight model and classifies hallucinations without ever touching
the generator. Across seven architectures from 0.5 billion to 9 billion parameters, we
consistently beat ReDeEP on RAGTruth AUC by 7.4 to 10.3 percentage points. Qwen2.5-7B
at an F1 of 0.717 is the first proxy-analyzer to surpass ReDeEP's token-level
threshold. All seven models land within a 2.3 percentage point AUC band, showing
detection quality saturates across scale. One of the most interesting results is that
our 3B LLaMA outperforms our 8B LLaMA on the primary benchmark, showing that bigger is
not always better even within the same model family. Architecture design is the stronger
quality driver. Because both RAGTruth and LLM-AggreFact draw from multiple generator
families, our results and conclusions are free from any generator-specific bias.

\end{document}